\documentclass[conference]{IEEEtran}
\IEEEoverridecommandlockouts

\usepackage{cite}
\usepackage{amsmath,amssymb,amsfonts}
\usepackage{algorithmic}
\usepackage{algorithm}
\usepackage{graphicx}
\usepackage{textcomp}
\usepackage{xcolor}
\usepackage{adjustbox}
\usepackage{multirow}
\usepackage{pifont}
\usepackage{enumitem}
\usepackage{tcolorbox}
\usepackage{booktabs}
\usepackage{subfig}
\usepackage{hyperref}
\newtcolorbox{AIbox}[1]{
    colback=gray!5!white,
    colframe=black!75!black,
    title=#1
}

\begin{document}
\newcommand{\stitle}[1]{\vspace*{0.5em}\noindent{\bf #1\/}}

\title{TCC-RAG: Temporal Causal Chain–Guided RAG for Temporal Knowledge Graph Question Answering
}

\author{
\IEEEauthorblockN{Xinxun Zhang\textsuperscript{1*}, Yuxiang Wang\textsuperscript{1*}, Shanshan Feng\textsuperscript{1}, Jun Zhang\textsuperscript{2}, Tieyun Qian\textsuperscript{1}, \\
Binhang Yuan\textsuperscript{3}, Kai Yin\textsuperscript{1$\dagger$}, Xiao Yan\textsuperscript{1}, Jiawei Jiang\textsuperscript{1,4$\dagger$}%
\thanks{\textsuperscript{*}Equal contribution. \textsuperscript{$\dagger$}Corresponding authors.}}

\IEEEauthorblockA{$^1$ School of Computer Science, Wuhan University \quad $^2$ Tencent Inc.}
\IEEEauthorblockA{$^3$ Hong Kong University of Science and Technology}
\IEEEauthorblockA{$^4$ School of Computer Science, Central China Normal University}

\IEEEauthorblockA{ \{{xxzhangstu,nai.yxwang,victor\_fengss,qty}\}@whu.edu.cn;  {neoxzhang}@tencent.com\\
{biyuan}@ust.hk;\{{yinkai,yanxiaosunny,jiawei.jiang}\}@whu.edu.cn
 }
}

\maketitle

\begin{abstract}
  Recent advances in large language models (LLMs) have spurred growing interest in temporal knowledge graph question answering (TKGQA). Most existing methods rely on semantic retrieval augmented generation (RAG), which suffers from fundamental limitations in complex temporal reasoning. On the one hand, semantic-based RAG methods struggle to capture the multi-hop causal knowledge required for complex temporal reasoning. On the other hand, these methods introduce excessive redundant knowledge, trapping LLMs in spurious reasoning chains that potentially result in erroneous answers. To address these problems, we propose a Temporal Causal Chain–guided RAG (TCC-RAG) to retrieve a concise yet sufficient set of causal knowledge to support complex temporal reasoning. Specifically, we reformulate the temporal knowledge graph into a multi-view causal cue graph, with causal cues inducing connections that explicitly encode temporal evolution and causal dynamics. Based on the causal cue graph, we construct a query-specific temporal causal chain to guide the retrieval process, enabling the model to focus on a concise set of causally relevant knowledge. Finally, we introduce a causal evidence filter to refine the retrieved knowledge, ensuring its causal consistency with the underlying temporal logic of the query. Experiments show that TCC-RAG achieves substantial performance gains of 10.72\% and 8.40\% on two TKGQA benchmarks, and delivers a remarkable 62.97\% improvement for multi-step temporal reasoning.

\end{abstract}


\section{Introduction}
Large Language Models (LLMs), such as GPT-4 \cite{achiam2023gpt} and the LLaMA series \cite{touvron2023llama}, have brought a paradigm shift to artificial intelligence by demonstrating strong capabilities in semantic understanding, reasoning, and generalization across a wide range of tasks \cite{li2025generation,yu2025survey,han2026micu,luo2025guiding,luo2026adamix,pmlr-v306-zheng26ae}. These advances have made LLMs increasingly attractive for question answering (QA), where models are expected to understand natural language queries and produce accurate, knowledge-grounded responses \cite{yu2025survey}. To improve factual reliability, Retrieval-Augmented Generation (RAG) \cite{fan2024RAGsurvey,zhao2024RAGsurvey2,gao2023retrieval} has become a widely adopted paradigm for QA systems. By retrieving relevant knowledge from external sources and incorporating it into the LLM context, RAG effectively supplements LLMs and improves factual grounding, particularly in knowledge-intensive reasoning tasks.

\begin{figure}[t]
\centering
\includegraphics[width=0.85\linewidth]{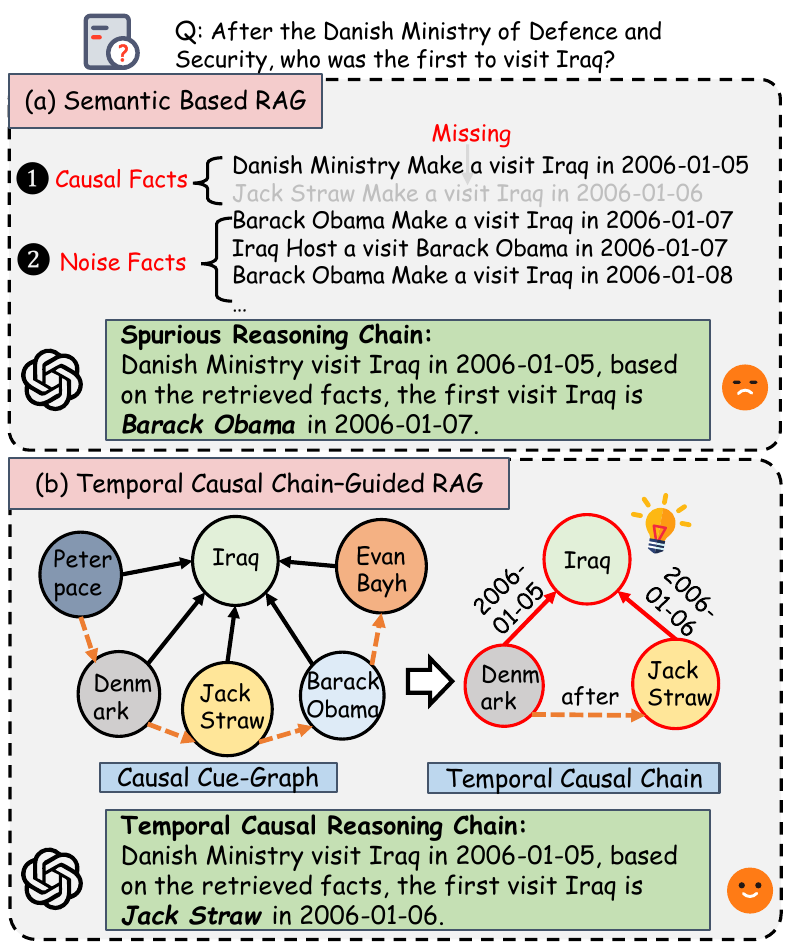}
\caption{The limitations of semantic-based RAG methods and design of our temporal causal chain-guided RAG approach.}
\label{fig_motivation}
\end{figure}

Temporal Knowledge Graph Question Answering (TKGQA) is an important QA task that requires reasoning over knowledge that evolves over time \cite{su2024temporal,chen2023multi}. Early TKGQA methods primarily relied on embedding-based approaches \cite{devlin2019bert,Lan2020ALBERT,mavromatis2022tempoqr}, which learn representations of temporal entities and relations to support inference over temporal knowledge graphs. More recently, with the rapid development of LLMs, researchers have begun to explore LLM-based TKGQA frameworks that leverage the reasoning ability of LLMs together with external temporal knowledge \cite{piryani2025suvery}. Most of these methods adopt semantic retrieval-augmented generation, where temporally relevant facts are retrieved based on semantic similarity and provided to the LLM for reasoning \cite{qian2024timer4,gong2025rtqa,gao2024two,chen-2024-ari,qianyihu2025timeagent}. However, such semantic RAG pipelines struggle to support complex multi-step temporal reasoning, where answering a query requires reconstructing temporally ordered reasoning chains across multiple events. In particular, existing methods suffer from two fundamental limitations in complex temporal reasoning scenarios, as illustrated in Figure~\ref{fig_motivation}(a).

\noindent \begin{itemize}[leftmargin=*]

\item[\ding{182}] \textit{Missing Causal Facts.}
Semantic-based RAG methods treat knowledge as isolated and unordered text units, ignoring the temporal evolution inherent in the data. Complex multi-step reasoning questions usually require multi-hop causal knowledge to reconstruct the correct temporal reasoning process. However, multi-hop causal knowledge that is causally essential but only partially relevant to the query semantics may be overlooked, preventing LLMs from capturing the true temporal causal chain in multi-step temporal reasoning scenarios. As illustrated in Figure \ref{fig_motivation}(a), for the multi-step temporal reasoning question
\textit{"After the Danish Ministry of Defence and Security, who was the first to visit Iraq?"}, semantic-based retrieval fails to retrieve the causally necessary fact
\textit{"Jack Straw made a visit to Iraq on 2006-01-06"} as it has limited semantic overlap with the query despite being causally essential for multi-step temporal reasoning.

\item[\ding{183}]\textit{Introducing Noise Facts.} Semantic-based RAG methods retrieve a large number of knowledge units through similarity matching, most of which are redundant, while only a small portion is truly causally relevant to the query. The presence of enormous noisy knowledge can lead LLMs to follow spurious reasoning chains, which may ultimately induce incorrect answers. As illustrated in Figure \ref{fig_motivation}(a), semantic-based RAG methods
return facts that are semantically related but causally irrelevant, such as
\textit{"Iraq hosted a visit by Barack Obama on 2006-01-07"}
and \textit{"Barack Obama made a visit to Iraq on 2006-01-08"}.
The presence of these two causally irrelevant facts about \textit{Barack Obama} may cause the LLM to follow spurious reasoning chains based on \textit{Barack Obama}'s visits when performing temporal reasoning, leading to incorrect responses.

\end{itemize}

\stitle{Our Solution: TCC-RAG.} To address the aforementioned limitations, we propose Temporal Causal Chain–guided Retrieval-Augmented Generation (TCC-RAG), a novel framework that retrieves a concise yet sufficient set of causal knowledge to support complex temporal reasoning. Specifically, we reformulate temporal knowledge facts into a multi-view causal cue graph, where temporal evolution is explicitly encoded as causal cues across one event graph and two trie-structured indices. This transformation exposes temporal evolution as explicit causal logic between events, capturing structural dynamics that plain text semantics cannot. Moreover, we introduce a temporal causal chain-guided retrieval strategy that progressively parses queries and retrieves only the causally relevant facts along the routed view of the causal cue graph. This design enables the retrieval process to focus on a concise yet sufficient causal knowledge, effectively avoiding the introduction of redundant knowledge that misleads LLMs into incorrect reasoning paths. Finally, to further ensure the causal consistency of the retrieved knowledge, we propose a causal evidence filter that distills the retrieved evidence into a minimal causally sufficient subset through two deterministic steps: equivalence-class deduplication and sufficiency-driven selection. The filter yields a purified evidence set that provides more faithful support for temporal reasoning.

We conduct a comprehensive evaluation of TCC-RAG on two widely used TKGQA benchmarks against over 10 baselines, covering diverse temporal reasoning scenarios.  TCC-RAG consistently outperforms all baselines, achieving overall improvements of 10.72\% and 8.40\% in Hits@1, respectively. Notably, it yields a substantial improvement of 62.97\% on complex multi-step temporal reasoning, demonstrating its strong capability in complex temporal reasoning. Ablation studies and case studies further validate the effectiveness of each component and the quality of the retrieved causal evidence. Efficiency analysis shows that TCC-RAG reduces token consumption by over 57\% compared with the strongest baseline, confirming its practical efficiency and scalability.

The main contributions of this work are as follows:

\begin{itemize}[leftmargin=*]
    \item We identify the limitations of existing semantic-based RAG methods for TKGQA, including the absence of causal facts and the introduction of enormous noisy facts, which hinder complex multi-step temporal reasoning.

    \item We propose {TCC-RAG}, which organizes temporal knowledge into a {multi-view causal cue graph} to explicitly encode temporal evolution dynamics as structured causal cues.

    \item We design a {temporal causal chain-guided retrieval strategy} with a {causal evidence filter} to retrieve and distill relevant facts into causal evidence for complex temporal reasoning.
\end{itemize}

\section{Related Work}

\stitle{Retrieval-Augmented Generation.}
Retrieval-Augmented Generation (RAG) enhances large language models by retrieving external knowledge to support reasoning and generation \cite{fan2024RAGsurvey,gao2023retrieval,zhao2024RAGsurvey2,huang2026dm}. A typical RAG framework consists of a retriever that identifies relevant documents or facts and a generator that produces answers conditioned on the retrieved evidence. For instance, \cite{chen2024nativerag} conduct a comprehensive benchmark of LLMs in RAG settings, revealing that retrieval noise and context length are key bottlenecks for faithful generation. HippoRAG \cite{jimenez2024hipporag} draws inspiration from human memory consolidation to build a long-term associative memory structure, enabling multi-hop retrieval by linking retrieved passages through a hippocampal-like index. These works highlight that the quality and structure of retrieved evidence are critical to downstream reasoning performance.

GraphRAG has attracted increasing attention due to its ability to model structured knowledge and capture multi-hop contextual dependencies \cite{edge2024graphrag,sun2024tog,mavromatis2025gnnrag,zeng2026pop}. Instead of retrieving independent text chunks, GraphRAG methods organize knowledge as graphs and perform structure-aware retrieval. For example, Think-on-Graph (ToG) \cite{sun2024tog} iteratively explores knowledge graph paths via LLM-guided beam search, retrieving multi-hop reasoning chains in a depth-first manner. GNN-RAG \cite{mavromatis2025gnnrag} combines graph neural networks with RAG by first extracting candidate answer entities through GNN reasoning and then retrieving shortest paths as evidence for LLM generation. Existing GraphRAG approaches can generally be categorized into subgraph retrieval methods \cite{li2025subgrapgrag} and path-based reasoning methods \cite{tan2025paths}, both of which aim to leverage graph structure to improve reasoning quality.
However, how GraphRAG can effectively support complex temporal reasoning, especially multi-step temporal reasoning, remains largely underexplored.

\stitle{Temporal Knowledge Graph Question Answering.}
Temporal Knowledge Graph Question Answering, a crucial paradigm for modeling evolving knowledge, has recently garnered significant research attention \cite{su2024temporal,chen2023multi}. Existing TKGQA methods can be broadly categorized into {embedding-based approaches} and {LLM-based approaches} \cite{piryani2025suvery}.

Embedding-based methods typically learn representations of temporal entities and relations and perform reasoning in the embedding space. CronQA \cite{saxena2021cronQA} encodes temporal constraints directly into entity and relation embeddings via a time-aware scoring function, enabling the model to answer questions anchored to specific timestamps. TempoQR \cite{mavromatis2022tempoqr} extends this by grounding question representations into the temporal knowledge graph through entity and time mention linking, achieving stronger alignment between query semantics and temporal facts. Other approaches rely on graph neural networks to model temporal relational structures \cite{jia2024faithfulgnn1,liu2023local,xue2024questiongnn3,sharma2023twirgcn}, capturing local and global structural dependencies within the knowledge graph. While effective on simpler temporal queries, these embedding-based methods generally struggle to generalize to complex multi-step temporal reasoning scenarios.

More recently, LLM-based TKGQA methods have emerged as a flexible alternative \cite{piryani2025suvery}. TimeR4 \cite{qian2024timer4} proposes a time-aware retrieval strategy that incorporates temporal signals into the retrieval process, providing LLMs with temporally filtered evidence for question answering. RTQA \cite{gong2025rtqa} introduces a recursive thinking mechanism that decomposes complex temporal questions into sub-questions and iteratively retrieves supporting facts, improving multi-step reasoning performance. Despite these advances, most existing LLM-based methods still rely on semantic similarity-based retrieval \cite{qian2024timer4,gong2025rtqa,chen-2024-ari}, which fails to capture the multi-hop causal knowledge required for complex temporal reasoning, motivating our causal chain-guided approach.

\section{Preliminaries}
\label{sec:preliminaries}
\stitle{Temporal Knowledge Graph.}
A temporal knowledge graph (TKG) is defined as a structured knowledge base that represents relational facts evolving over time. Formally, a TKG is denoted as: $\mathcal{G} = (\mathcal{E}, \mathcal{R}, \mathcal{T}, \mathcal{F}),$ where $\mathcal{E}$ denotes the entity set, $\mathcal{R}$ the relation set, $\mathcal{T}$ the temporal domain, and $\mathcal{F}$ the complete temporal fact set. Each temporal fact is represented as a quadruple $(h, r, t, \tau) \in \mathcal{F},$ where $h, t \in \mathcal{E}$ are the head and tail entities, $r \in \mathcal{R}$ is the relation type, and $\tau \in \mathcal{T}$ denotes the temporal annotation indicating when the relation holds. The temporal annotation $\tau$ can be a discrete timestamp, a time interval, or an ordered temporal index.

\stitle{Temporal Knowledge Graph Question Answering.}
Given a temporal knowledge graph $\mathcal{G}$ and a natural language question $\mathcal{Q}$, the goal of temporal knowledge graph question answering (TKGQA) is to infer the correct answer $\mathcal{A}$ grounded in $\mathcal{G}$ while respecting temporal constraints and event evolution patterns. Formally, the task is to learn a mapping: $f: ( \mathcal{Q}, \mathcal{G}) \rightarrow \mathcal{A},$ where $\mathcal{A} \in \mathcal{E}$, $\mathcal{A} \in \mathcal{T}$, or $\mathcal{A} \in \mathcal{E} \times \mathcal{T}$, depending on the question type.

\stitle{Complex Temporal Reasoning.}
Unlike static KGQA, TKGQA often requires multi-step reasoning over time. Specifically, the answer $a$ may depend on a sequence of temporally ordered facts:
$\pi = { (h_1, r_1, t_1, \tau_1), \dots, (h_n, r_n, t_n, \tau_n) },$
where the timestamps satisfy temporal ordering constraints:
$\tau_1 \prec \tau_2 \prec \dots \prec \tau_n,$
and collectively provide sufficient evidence to derive the answer:
$\pi \models a \mid \mathcal{Q}.$

\stitle{Temporal Causal Cue.}
A \textit{causal cue} is a query-conditioned temporal dependency signal that constrains how query-relevant events are ordered or selected. We distinguish two types of cues. The first type is the \textit{ordering cue} $c \in \{\texttt{before}, \texttt{after}, \texttt{same}\}$, which constrains the temporal relation between an event and a boundary timestamp $\tau_b$. Given a query $\mathcal{Q}$ with ordering cue $\hat{c}$, an event $e = (h, r, t, \tau)$ is \textit{cue-consistent} with $\mathcal{Q}$ if its timestamp $\tau$ satisfies the constraint imposed by $\hat{c}$ relative to $\tau_b$:
\begin{equation}
\tau \prec_{\hat{c}} \tau_b \;\Leftrightarrow\;
\begin{cases}
\tau < \tau_b, & \hat{c} = \texttt{before}, \\
\tau > \tau_b, & \hat{c} = \texttt{after},  \\
\tau = \tau_b, & \hat{c} = \texttt{same}.
\end{cases}
\end{equation}
The second type is the \textit{selection cue} $c \in \{\texttt{first}, \texttt{last}\}$, which selects the boundary event from a temporally ordered set of cue-consistent events. Specifically, \texttt{first} selects the earliest event and \texttt{last} selects the latest one. Selection cues are typically composed with ordering cues in multi-step queries, for example \emph{``Before Mexico, which country did China visit last?''}, where \texttt{before} first restricts the candidate events and \texttt{last} then selects the answer among them.

In this work, the term causal is used in an operational sense for temporal question answering rather than in the intervention-based sense of causal inference. We regard temporally ordered event dependencies that are relevant to a query as causal signals for reasoning.

\begin{figure*}[htbp]
\centering
\includegraphics[width=0.99\linewidth]{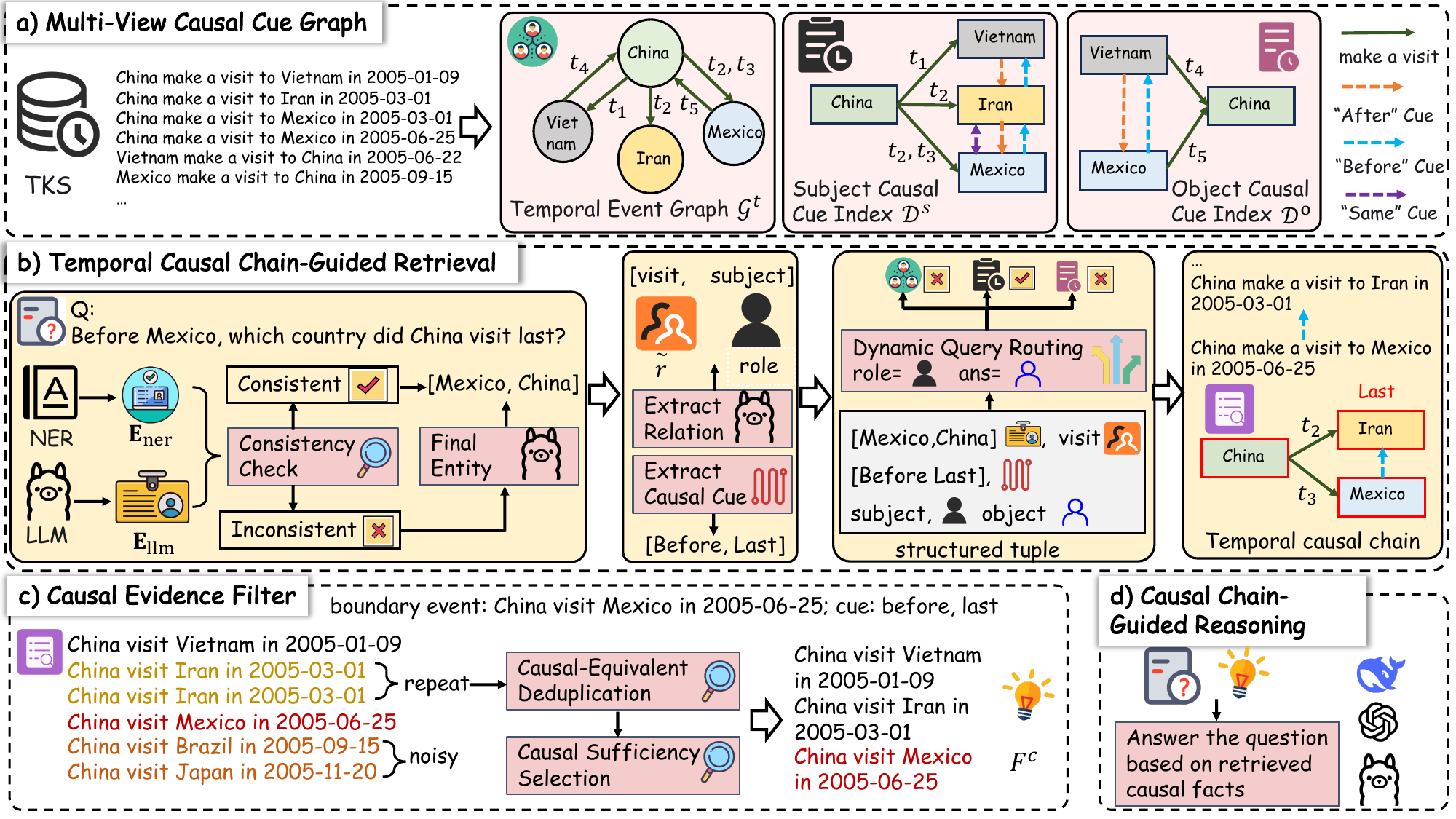}
\caption{The framework of TCC-RAG. (a) Temporal facts are organized into a Temporal Event Graph (TEG) and two temporally sorted indices: the Subject Causal Cue Index (SCCI) and Object Causal Cue Index (OCCI). (b) NER and LLM extraction ground query entities through matching and consistency checking. Relation, cue, and role parsing then guide view routing and temporal chain retrieval. (c) The Causal Evidence Filter removes equivalent duplicates and selects a sufficient subset of query-consistent facts. (d) The LLM reasons over the filtered evidence to generate answers. When structured retrieval yields no facts, semantic retrieval provides fallback candidates that also pass through the filter.}
\label{fig_model}
\end{figure*}

\section{Methodology}
TCC-RAG consists of three core components. First, we construct a multi-view causal cue graph to explicitly encode temporal causal relations among events, transforming implicit temporal information into structured causal cues organized over a graph and two indices. Second, we design a query-aware temporal causal chain retrieval module that progressively parses the query and traces causally relevant evidence over the routed view, ensuring that only essential causal facts are retrieved. Finally, we introduce a causal evidence filter that refines the retrieved evidence into a minimal causally sufficient subset through equivalence-class deduplication and sufficiency-driven selection. Together, these components form a unified, training-free, and plug-and-play framework that effectively enhances LLM-based complex temporal reasoning. The overall architecture of TCC-RAG is shown in Figure \ref{fig_model}.

\subsection{Multi-View Causal Cue Graph Construction}
\label{sec:causal-cue-graph}
Different temporal queries exhibit heterogeneous logical structures and require varying reasoning depths. Simple queries that involve a known entity pair only require direct event lookup, while complex multi-step queries anchored on a single entity demand traversal across temporally ordered events. A single monolithic structure cannot serve both efficiently. To this end, we design the causal cue graph as three complementary views: a \textit{Temporal Event Graph} (TEG) for paired-entity queries, a \textit{Subject Causal Cue Index} (SCCI) for subject-anchored multi-step queries, and an \textit{Object Causal Cue Index} (OCCI) for object-anchored multi-step queries. The model dynamically routes each query to the most suitable view based on the role of its grounded entities, achieving both precision and efficiency across diverse query patterns.

\stitle{Temporal Event Graph.}
TEG targets paired-entity queries that ask for events directly connecting two known entities, e.g., \textit{``When did China negotiate with Iraq?''}. For such queries, retrieval reduces to a direct edge lookup between two nodes, which is naturally captured by a graph. Formally, we construct TEG as a multi-relational temporal graph: $\mathcal{G}^{t} = (\mathcal{E}, \mathcal{F}),$
where $\mathcal{E}$ is the entity node set (as defined in Section~\ref{sec:preliminaries}), and each edge $(h, r, t, \tau) \in \mathcal{F}$ encodes a temporal fact linking head $h$ to tail $t$ via relation $r$ at timestamp $\tau$. Given the grounded subject--object pair $(\hat{e}_s,\hat{e}_o)$, retrieval scans directed edges from $\hat{e}_s$ to $\hat{e}_o$ and retains those matching the query relation, costing $\mathcal{O}(\mathrm{deg}(\hat{e}_s,\hat{e}_o))$, where $\mathrm{deg}$ counts these directed edges.

\stitle{Subject Causal Cue Index.}
A more challenging case arises in multi-step queries anchored on a subject entity, such as \textit{``After Mexico, which country did China visit?''}. The answer requires traversing all events initiated by the subject under the same relation, ordered along the temporal dimension. A naive solution is to materialize a causal cue edge between every pair of temporally related events sharing the same $(subject, relation)$ key. However, this incurs $\mathcal{O}(d^{2})$ storage cost per key, where $d$ denotes the number of events per $(subject, relation)$ pair, becoming prohibitive in dense entity neighborhoods.

We observe that traversing along causal cue edges is functionally equivalent to scanning a list of events sorted by timestamp. Causal cues encode the relative temporal order between events, and this order is fully recoverable from the position of each event in the sorted list. Building on this insight, we design SCCI as an inverted index that encodes causal cues implicitly through temporal ordering, rather than as physically materialized edges.
\begin{table}[t]
\centering
\caption{Complexity of the multi-view causal cue graph. $|\mathcal{F}|$: total facts; $d_{\max}$: maximum posting list length; $k$: number of cue-consistent events per query. \emph{Per-query} measures index retrieval with grounded keys and a resolved temporal boundary (see Section~\ref{sec:complexity}).}
\label{tab:causal-cue-complexity}
\small
\begin{tabular}{lccc}
\toprule
Structure & Storage & Construction & Per-query \\
\midrule
TEG\ ($\mathcal{G}^{t}$)  & $\mathcal{O}(|\mathcal{F}|)$ & $\mathcal{O}(|\mathcal{F}|)$ & $\mathcal{O}(\mathrm{deg})$ \\
SCCI\ ($\mathcal{D}^{s}$) & $\mathcal{O}(|\mathcal{F}|)$ & $\mathcal{O}(|\mathcal{F}| \log d_{\max})$ & $\mathcal{O}(\log d_{\max}{+}k)$ \\
OCCI\ ($\mathcal{D}^{o}$) & $\mathcal{O}(|\mathcal{F}|)$ & $\mathcal{O}(|\mathcal{F}| \log d_{\max})$ & $\mathcal{O}(\log d_{\max}{+}k)$ \\
\bottomrule
\end{tabular}
\end{table}
Concretely, SCCI is a hash map keyed by $(subject, relation)$ pairs, with each value storing a temporally ordered posting list:
\begin{equation}
\mathcal{D}^{s}: (u, r) \mapsto \big[\, e_1, e_2, \dots, e_{d}\,\big],
\quad \tau_{e_1} \le \tau_{e_2} \le \dots \le \tau_{e_d}.
\end{equation}
The causal cues among events in the posting list are recovered on-the-fly: if $\tau_{e_i}<\tau_{e_j}$, then $\texttt{before}(e_i,e_j)$ and $\texttt{after}(e_j,e_i)$ hold. If $\tau_{e_i}=\tau_{e_j}$, only $\texttt{same}(e_i,e_j)$ holds. Each predicate describes its first event relative to its second. List position alone does not imply strict temporal precedence. This implicit encoding preserves the full causal cue semantics defined in Section~\ref{sec:preliminaries}, while reducing storage from $\mathcal{O}(d^{2})$ to $\mathcal{O}(d)$ per key.

The SCCI design delivers two efficiency benefits over a fully materialized causal cue graph. First, storage is reduced from $\mathcal{O}(d^2)$ to $\mathcal{O}(d)$ per key, yielding $\mathcal{O}(|\mathcal{F}|)$ total. Second, a hash lookup locates the posting list, binary search locates the temporal range, and a scan returns its $k$ cue-consistent events, costing $\mathcal{O}(\log d_{\max}+k)$. Crucially, this efficiency comes without sacrificing causal cue semantics: every retrieval result reachable via materialized edges remains reachable via posting list scan.

\stitle{Object Causal Cue Index.}
Symmetric to SCCI, OCCI handles multi-step queries anchored on an object entity, e.g., \textit{``Before the United States, who visited Iraq?''}. Following the same design rationale, we construct OCCI as a hash map keyed by $(relation, object)$ pairs:
\begin{equation}
\mathcal{D}^{o}: (r, u) \mapsto \big[\, e_1, e_2, \dots, e_{d}\,\big],
\quad \tau_{e_1} \le \tau_{e_2} \le \dots \le \tau_{e_d}.
\end{equation}
The causal cue semantics, storage cost $\mathcal{O}(|\mathcal{F}|)$, and per-query retrieval cost $\mathcal{O}(\log d_{\max}+k)$ are identical to those of SCCI, providing the same efficiency guarantees on the object-anchored side.

\stitle{Summary.}
We denote the multi-view structure by $\mathcal{S}=(\mathcal{G}^{t},\mathcal{D}^{s},\mathcal{D}^{o})$, distinct from the original TKG $\mathcal{G}$. Construction populates the three views in one pass over $\mathcal{F}$ and then sorts the posting lists, taking $\mathcal{O}(|\mathcal{F}|\log d_{\max})$ time and $\mathcal{O}(|\mathcal{F}|)$ storage. Table~\ref{tab:causal-cue-complexity} summarizes the view-specific retrieval costs. Together, the views preserve causal cue semantics through direct edges and implicit temporal ordering.

\subsection{Temporal Causal Chain-Guided Retrieval}\label{sec:retrieval}
Given the multi-view causal cue graph, the central challenge is to ground a natural-language query $\mathcal{Q}$ into the appropriate index keys and extract the events that participate in the underlying temporal causal chain. We address this challenge through a progressive query parsing pipeline that decomposes parsing into three stages of decreasing difficulty: hybrid entity grounding, context-aware relation and cue parsing, and dynamic query routing. Each stage takes the signals resolved by the previous one as additional context, so that the search space narrows along the pipeline and progressively lighter machinery suffices for the remaining decisions. The structured output is finally resolved into a temporal causal chain by a chain construction procedure over the routed view.

\begin{algorithm}[t]
\small

\caption{Temporal Causal Chain-Guided Retrieval}
\label{alg:retrieval}
\begin{algorithmic}[1]
\REQUIRE Query $\mathcal{Q}$, views $\mathcal{S}=(\mathcal{G}^{t},\mathcal{D}^{s},\mathcal{D}^{o})$, KG vocabulary $\mathbf{E}_{\mathrm{cand}}$, relations $\mathcal{R}$.
\ENSURE Evidence $F^{c}$ for answer generation.
\STATE \textit{// Hybrid Entity Grounding}
\STATE $M_{\mathrm{ner}} \leftarrow \mathrm{NER}(\mathcal{Q})$
\STATE $M_{\mathrm{llm}} \leftarrow \mathrm{LLM}(\mathcal{Q};\mathrm{prompt}_{\mathrm{ent}})$
\STATE Match both mention sets to $\mathbf{E}_{\mathrm{cand}}$ using Eq.~\eqref{eq:entity-grounding}
\STATE Obtain $\hat{\mathbf{E}}$ by agreement or arbitration in Eq.~\eqref{eq:entity-arbitration}
\STATE \textit{// Relation, Cue, and Role Parsing}
\STATE $\hat{c} \leftarrow \mathrm{RuleMatch}(\mathcal{Q}; \mathcal{P})$
\STATE $(\tilde{r},\mathrm{role},z_{\mathrm{ref}}) \leftarrow \mathrm{LLM}(\mathcal{Q},\hat{\mathbf{E}};\mathrm{prompt}_{\mathrm{rel}})$
\STATE $r^{*} \leftarrow \arg\max_{r \in \mathcal{R}} \cos(\phi(\tilde{r}), \phi(r))$
\STATE Infer $\mathrm{ans}(\mathcal{Q})$ and filled slots from $\mathcal{Q}$ and $\mathrm{role}$
\STATE Resolve $\tau_b$ from $\mathcal{Q}$ and $z_{\mathrm{ref}}$, using $\bot$ if absent or unresolved
\STATE \textit{// Route to One View and Construct the Chain}
\STATE $F_Q \leftarrow \emptyset$
\IF{$\mathrm{pair}(\mathcal{Q})$}
  \STATE $F_Q \leftarrow \mathrm{DirectedEdges}(\mathcal{G}^{t},\hat e_s,r^{*},\hat e_o)$
\ELSIF{$\mathrm{role}(\hat e)=\mathrm{subject}$ and $\mathrm{ans}(\mathcal{Q})=\mathrm{object}$}
  \STATE $F_Q \leftarrow \mathcal{D}^{s}[(\hat e,r^{*})]$
\ELSIF{$\mathrm{role}(\hat e)=\mathrm{object}$ and $\mathrm{ans}(\mathcal{Q})=\mathrm{subject}$}
  \STATE $F_Q \leftarrow \mathcal{D}^{o}[(r^{*},\hat e)]$
\ENDIF
\STATE $F_Q \leftarrow \mathrm{TemporalSelect}(F_Q,\hat c,\tau_b)$
\IF{$F_Q=\emptyset$}
  \STATE $F_Q \leftarrow \mathrm{SemanticRetrieve}(\mathcal{Q})$
\ENDIF
\STATE Order $F_Q$ by ascending timestamp
\STATE $F^{c} \leftarrow \mathrm{CEF}(F_Q,\mathcal{Q})$ \COMMENT{Section~\ref{sec:csef}}
\STATE \textbf{return} $F^{c}$
\end{algorithmic}
\end{algorithm}

\stitle{Hybrid Entity Grounding.}
Entity grounding is the most challenging stage of parsing because it operates on the raw query with no prior context. Real-world temporal queries frequently involve aliases, composite references, and role descriptors, all of which may map to canonical KG entities in non-obvious ways. A single extractor, whether a NER model or an LLM, may fail to handle such surface-form variability reliably. To address this, we adopt a dual-path extraction design with mutual consistency verification.

For query $\mathcal{Q}$, NER and the LLM independently extract mention sets $M_{\mathrm{ner}}$ and $M_{\mathrm{llm}}$. Both are matched against the KG entity vocabulary $\mathbf{E}_{\mathrm{cand}}=\mathcal{E}$, not a query-derived candidate set:
{
\begin{equation}
\label{eq:entity-grounding}
\begin{aligned}
\mathbf{E}_{\mathrm{ner}} &= \mathcal{M}(M_{\mathrm{ner}},\mathbf{E}_{\mathrm{cand}}), \\
\mathbf{E}_{\mathrm{llm}} &= \mathcal{M}(M_{\mathrm{llm}},\mathbf{E}_{\mathrm{cand}}).
\end{aligned}
\end{equation}
}
Here $\mathbf{E}_{\mathrm{ner}}$ and $\mathbf{E}_{\mathrm{llm}}$ contain matched KG entities, and $\mathcal{M}(\cdot)$ is a hybrid matching function. For each extracted mention, $\mathcal{M}(\cdot)$ first applies semantic matching via cosine similarity in a pretrained embedding space. If the maximum similarity exceeds threshold $\theta_1$, the semantic result is adopted. Otherwise, it falls back to string-based fuzzy matching.

We then apply a consistency check to produce the final grounded entity set $\hat{\mathbf{E}}$:
\begin{equation}
\label{eq:entity-arbitration}
\hat{\mathbf{E}} =
\begin{cases}
\mathbf{E}_{\mathrm{ner}}, & \text{if } \mathbf{E}_{\mathrm{ner}} = \mathbf{E}_{\mathrm{llm}}, \\
\mathrm{LLM}(\mathcal{Q},\, \mathbf{E}_{\mathrm{ner}},\, \mathbf{E}_{\mathrm{llm}};\, \mathrm{prompt}_{\mathrm{com}}), & \text{otherwise.}
\end{cases}
\end{equation}
When the two paths agree, the shared result is directly adopted. When they disagree, the LLM is prompted with both candidates and the original query to resolve the conflict and select the correct entity set. This dual-path design enables self-correction without labeled data or fine-tuning, improving grounding robustness across diverse and linguistically complex temporal queries.

\stitle{Context-Aware Relation and Cue Parsing.}
With entities grounded, the second stage extracts two predicate-level signals, the temporal cue $\hat{c}$ and the canonical relation $r^{*}$, which together determine the target posting list within $\mathcal{D}^{s}$ or $\mathcal{D}^{o}$, or the target edge within $\mathcal{G}^{t}$. Because cues and relations exhibit different parsing complexities, we adopt different extraction strategies for the two.

For temporal cues, we observe that nearly all temporal queries express their cue through a small closed set of operator words such as \texttt{before}, \texttt{after}, \texttt{first}, and \texttt{last}, often in regular surface patterns. We therefore extract $\hat{c}$ via deterministic pattern matching on the query surface:
\begin{equation}
\hat{c} = \mathrm{RuleMatch}(\mathcal{Q}; \mathcal{P}),
\end{equation}
where $\mathcal{P}$ is a small set of cue-pattern rules.

For relations, we leverage the LLM's predicate-parsing capability conditioned on the grounded entities $\hat{\mathbf{E}}$. The LLM extracts a relation phrase $\tilde{r}$, subject/object role assignments, and a temporal reference $z_{\mathrm{ref}}$ when present:
{
\begin{equation}
(\tilde{r},\mathrm{role},z_{\mathrm{ref}}) = \mathrm{LLM}(\mathcal{Q},\hat{\mathbf{E}};\mathrm{prompt}_{\mathrm{rel}}),
\end{equation}
}
where $\mathrm{prompt}_{\mathrm{rel}}$ denotes the relation-extraction template. The extracted phrase $\tilde{r}$ is then grounded to a canonical KG relation via embedding similarity over the relation set $\mathcal{R}$:
\begin{equation}
r^{*} = \arg\max_{r \in \mathcal{R}} \cos(\phi(\tilde{r}), \phi(r)).
\end{equation}
The answer slot $\mathrm{ans}(\mathcal{Q})$ follows from the interrogative and missing argument in the parsed subject--relation--object structure. For entity answers, the filled subject or object serves as the traversal anchor $\hat e$. An entity or date associated with the temporal cue provides the temporal reference $z_{\mathrm{ref}}$. For time answers, $\hat e_s$ and $\hat e_o$ denote the filled subject and object. These roles, together with $r^{*}$ and $\hat c$, determine routing and temporal selection.

\stitle{Dynamic Query Routing.}
With the parsed signals in hand, the third stage dynamically maps each query to the appropriate causal cue graph view based on the query structure. In particular, a multi-step subject-anchored or object-anchored query may still contain two grounded entities: one serves as the traversal anchor, while the other only provides the temporal reference point that determines $\tau_b$. We therefore route by the filled argument slot and the answer slot. Let $\mathrm{ans}(\mathcal{Q}) \in \{\mathrm{subject}, \mathrm{object}, \mathrm{time}\}$ denote the slot being asked, and let $\mathrm{pair}(\mathcal{Q})$ indicate that both subject and object slots are filled and the query asks for a direct temporal fact. The routing function $\rho: \mathcal{Q} \mapsto \{\mathcal{G}^{t}, \mathcal{D}^{s}, \mathcal{D}^{o}\}$ is defined as
\begin{equation}
\rho(\mathcal{Q}) =
\begin{cases}
\mathcal{G}^{t}, & \mathrm{pair}(\mathcal{Q})=\mathrm{true}, \\
\mathcal{D}^{s}, & \mathrm{role}(\hat{e})=\mathrm{subject} \;\wedge\; \mathrm{ans}(\mathcal{Q})=\mathrm{object}, \\
\mathcal{D}^{o}, & \mathrm{role}(\hat{e})=\mathrm{object} \;\wedge\; \mathrm{ans}(\mathcal{Q})=\mathrm{subject}.
\end{cases}
\end{equation}
For example, in the query \textit{``Before Mexico, which country did China visit last?''}, both \textit{China} and \textit{Mexico} are grounded. However, \textit{China} is the subject anchor and \textit{Mexico} only determines the boundary timestamp, so the query is routed to the SCCI dictionary $\mathcal{D}^{s}$ rather than to TEG. The three branches are mutually exclusive for the supported query patterns. Reference-event lookup supplies a boundary, not a second answer traversal. The executor does not perform arbitrary cross-view, multi-relation joins.

\stitle{Causal Chain Construction.}
After routing, we construct the temporal causal chain by retrieving cue-consistent events from the routed view. The construction proceeds along two paths corresponding to the structural property of the view, followed by a unified temporal selection step.

If the query is routed to TEG, retrieval uses the known subject $\hat e_s$ and object $\hat e_o$. The query-specific candidate set $F_Q$ contains directed edges matching $r^{*}$:
{
\begin{equation}
F_Q = \{(h,r,t,\tau)\in\mathcal{F}\mid h=\hat e_s,\;t=\hat e_o,\;r=r^{*}\}.
\end{equation}
}

For SCCI or OCCI, we locate the posting list $L$ keyed by $(\hat e,r^{*})$ or $(r^{*},\hat e)$, respectively. An explicit date directly supplies $\tau_b$. Otherwise, TEG retrieves the reference event connecting the anchor and reference entity under $r^{*}$. Its timestamp supplies $\tau_b$. With the boundary resolved, binary search followed by a range scan selects cue-consistent events:
{
\begin{equation}
F_Q = \{e\in L\mid \tau_e\prec_{\hat c}\tau_b\},
\end{equation}
}
where $\prec_{\hat c}$ is defined in Section~\ref{sec:preliminaries}. We use $\tau_b=\bot$ when the boundary is absent or unresolved. Without an ordering constraint, the full list is eligible for selection. A missing key or an unresolved required boundary yields $F_Q=\emptyset$.

The unified \textsc{TemporalSelect} operation applies these temporal constraints, then retains the earliest or latest event(s) for \texttt{first}/\texttt{last}, if requested. Thus, compound cues such as \texttt{before-last} first restrict the range and then select its temporal extreme. Queries without temporal restrictions retain all relation-matched candidates. The selected events in $F_Q$ are arranged in temporal order to form the temporal causal chain
\begin{equation}
\pi = e_1 \xrightarrow{\hat{c}} e_2 \xrightarrow{\hat{c}} \cdots \xrightarrow{\hat{c}} e_k,
\end{equation}
which serves as the structured causal evidence for downstream answer generation.

Parsing uses two LLM calls (mention extraction and predicate parsing), plus one arbitration call only when the grounding paths disagree. When structured retrieval returns $F_Q=\emptyset$, the retrieval pipeline falls back to semantic similarity search to obtain alternative candidates. Grounding errors, relation mismatches, or missing facts can cause such empty results. Routing alone does not trigger fallback. Candidates from either retrieval path are passed to CEF before answer generation, as summarized in Algorithm~\ref{alg:retrieval}.

\subsection{Causal Evidence Filter}\label{sec:csef}
Retrieved evidence can contain redundant or temporally irrelevant facts. CEF processes the query-specific candidates $F_Q$ from either structured retrieval or semantic fallback in two deterministic steps, producing deduplicated evidence $F^{d}$ and final filtered evidence $F^{c}$. Already concise, cue-consistent evidence can pass through unchanged, while redundant or cue-inconsistent candidates are refined before answer generation.

\stitle{Causal-Equivalent Deduplication.}
The first kind of noise comes from facts that are structurally distinct yet carry identical causal information under the query. We capture this via a query-conditioned causal signature:
\begin{equation}
\sigma(e \mid \mathcal{Q}) = \big(h, r, t, \kappa(\tau, \hat{c}, \tau_b)\big),
\end{equation}
where $\kappa(\tau, \hat{c}, \tau_b)$ maps the timestamp $\tau$ to a discrete causal token. Let $c_{\kappa}$ be the ordering component of $\hat c$ if present, otherwise its selection component (or $\bot$ if neither exists). Compound cues use the ordering component for normalization and retain the selection component for sufficiency selection. Let $D_b(\tau)$ be $(\tau-\tau_b).{\rm days}$ for a resolved boundary, or $\tau.{\rm toordinal}()$ otherwise. Then
{
\begin{equation}
\kappa(\tau, \hat{c}, \tau_b) =
\begin{cases}
\lfloor D_b(\tau)/\delta \rfloor, & c_{\kappa}\in\{\texttt{before},\texttt{after}\}, \\
{\rm rank}(\tau;c_{\kappa}), & c_{\kappa}\in\{\texttt{first},\texttt{last}\}, \\
\tau.{\rm toordinal}(), & \text{otherwise}.
\end{cases}
\end{equation}
}
where $\delta$ is the temporal bucket width (set to 1 day, aligned with the finest granularity of MULTITQ), and ${\rm rank}(\tau;c_{\kappa})$ is the ordinal position within the $(h,r,t)$ group, sorted ascending for \texttt{first} or descending for \texttt{last}. Thus, range cues use day buckets, while \texttt{same} and absent cues preserve day-level identity. Two facts are causally equivalent under $\mathcal{Q}$ if and only if they share the same signature $\sigma$. We partition $F_Q$ into equivalence classes under $\sigma$ and retain the earliest-timestamp fact as the canonical representative of each class, yielding the deduplicated set $F^{d}$. This stage runs in $\mathcal{O}(|F_Q|)$ via hash-based grouping and is information-lossless: every fact in $F_Q$ has an equivalent representative in $F^{d}$.

\stitle{Causal Sufficiency Selection.}
The second kind of noise comes from facts that fail to participate in the causal chain required by the query. Eligible facts satisfy the ordering component when $\tau_b$ is resolved. When it is unresolved, CEF skips boundary-dependent tests to avoid unsupported pruning of fallback evidence. We select a minimal subset $F^{c} \subseteq F^{d}$ by a greedy procedure that enforces three conditions in order:

\noindent\textbf{(i) Boundary preservation.} For a \texttt{first}/\texttt{last} selection component, including compound cues, the algorithm seeds $F^{c}$ with the earliest/latest eligible fact in $F^{d}$. For \texttt{same} with a resolved $\tau_b$, it retains all facts on the anchor day. This ensures that the pivotal boundary event anchoring the causal chain is always retained.

\begin{figure}[t]
\begin{AIbox}{Prompt for final answer generation}
{\small\sffamily

\textbf{Instruction:}

You are a {temporal knowledge question answering assistant}.
Given a list of Historical facts and a Question, return {ONLY} a JSON object in this exact format:

\begin{verbatim}
{"answers": [...]}
\end{verbatim}

\textbf{Rules:}

\begin{itemize}

\item Return all possible answers and do {NOT} provide explanations.
\item Every returned entity must appear {EXACTLY} as written in the Historical facts.
\item Dates must follow ISO format (YYYY-MM-DD). If the question asks for:
\begin{itemize}
  \item year → return YYYY
  \item month → return YYYY-MM
  \item day → return YYYY-MM-DD
\end{itemize}

\end{itemize}

}

\end{AIbox}
\caption{The prompt template for causal chain-guided temporal reasoning based answer generation.}
\label{fig:prompt_reason}
\end{figure}

\noindent\textbf{(ii) Cue coverage.} The algorithm adds the canonical representative of each eligible class in $F^{d}$, using the ordering component of $\hat c$ when present (i.e., $\tau<\tau_b$ for \texttt{before}, $\tau>\tau_b$ for \texttt{after}, or the same day for \texttt{same}, with a resolved $\tau_b$). Classes that are cue-inconsistent are discarded. This step ensures that every distinct causal pattern satisfying the query's temporal constraint is represented in $F^{c}$.

\noindent\textbf{(iii) Chain connectivity.} With $F^{c}$ ordered by ascending timestamp, the algorithm verifies that consecutive facts share the grounded relation $r^{*}$ and respect the temporal direction implied by $\hat{c}$ (i.e., timestamps are non-decreasing). If this condition is violated, indicating that the selected facts do not form a coherent causal chain, the algorithm safely falls back to $F^{d}$ to avoid over-pruning.

Preserving the chronological order supplied by chain construction, selection and connectivity checks take $\mathcal{O}(|F^{d}|)$ time. Together with hash-based deduplication, CEF takes $\mathcal{O}(|F_Q|)$ time and produces filtered evidence $F^{c}$.

\subsection{Causal Chain-Guided Reasoning}\label{sec:reasoning}
With the evidence set $F^{c}$ in hand, we prompt the LLM to perform temporal reasoning and generate the final answer. Specifically, $F^{c}$ is serialized as a structured fact list and provided to the LLM alongside the original query $\mathcal{Q}$, enabling it to follow the temporal causal chain to derive the answer. The prompt template used for answer generation is shown in Figure~\ref{fig:prompt_reason}. Compared with semantic-based RAG methods, which feed a large number of loosely relevant facts into the LLM, TCC-RAG supplies a concise yet causally sufficient evidence set. This precision reduces reasoning noise and allows the LLM to focus on the true causal chain, yielding more accurate answers with substantially fewer facts.

\subsection{Complexity Analysis}\label{sec:complexity}

\stitle{Time Complexity.}
Offline construction costs $\mathcal{O}(|\mathcal{F}|\log d_{\max})$, dominated by posting-list sorting. Query parsing uses two LLM calls, plus one optional grounding-arbitration call. Answer generation adds one further call. Entity and relation matching depend on their vocabulary sizes and are separate from index retrieval. With grounded keys and a resolved boundary, TEG lookup costs $\mathcal{O}(\mathrm{deg}(\hat e_s,\hat e_o))$, while SCCI/OCCI range retrieval costs $\mathcal{O}(\log d_{\max}+k)$. A reference-event boundary additionally requires a TEG edge lookup. Ordering initially unsorted candidates costs $\mathcal{O}(|F_Q|\log |F_Q|)$. Posting-list results already inherit temporal order. CEF then costs $\mathcal{O}(|F_Q|)$. Table~\ref{tab:causal-cue-complexity} reports index retrieval alone, excluding parsing, evidence filtering, generation, and the semantic backend used only for fallback. Thus, structured retrieval accesses query-relevant neighborhoods rather than scanning all $|\mathcal{F}|$ facts.

\stitle{Space Complexity.}
The three structured views require $\mathcal{O}(|\mathcal{F}|)$ storage: each fact appears once in TEG and once in each of SCCI and OCCI. These views require no fact-level dense vectors. Entity/relation embedding caches and the semantic fallback backend are additional storage, separate from this structural bound.

\subsection{Limitations}\label{sec:limitations}
\noindent\textbf{Entity Grounding and Query Parsing Noise.} Ambiguous entity mentions or misaligned relations can produce empty or incorrect structured evidence even when the query is routed to a view. Semantic retrieval provides a safety net when the structured result is empty, but does not guarantee detection of nonempty, incorrect evidence. Robust grounding and relation disambiguation remain important challenges.

\noindent\textbf{Sparse Knowledge Graph.} The SCCI/OCCI indices gain efficiency when multiple facts share the same $(subject, relation)$ or $(relation, object)$ key. On sparse KGs, most posting lists contain a single entry. In such cases, multi-step causal chain retrieval degenerates to a single fact lookup, and the multi-view design provides limited benefit. Adapting TCC-RAG to sparse KG settings is left to future work.

\begin{table}[t]
\caption{Statistics of question categories in MULTITQ.}
\label{tab:multiTQ_stats}
\centering
\small
\begin{tabular}{lcccc}
\toprule
\multicolumn{2}{c}{\textbf{Category}} & \textbf{Train} & \textbf{Dev} & \textbf{Test} \\
\midrule
\multirow{3}{*}{Single}
  & Equal & 135,890 & 18,983 & 17,311 \\
  & Before/After & 75,340 & 11,655 & 11,073 \\
  & First/Last & 72,252 & 11,097 & 10,480 \\
\cmidrule(lr){1-5}
\multirow{3}{*}{Multiple}
  & Equal Multi & 16,893 & 3,213 & 3,207 \\
  & After First & 43,305 & 6,499 & 6,266 \\
  & Before Last & 43,107 & 6,532 & 6,247 \\
\cmidrule(lr){1-5}
& Total & 386,787 & 57,979 & 54,584 \\
\bottomrule
\end{tabular}
\end{table}

\begin{table}[t]
\caption{Statistics of question categories in TimeQuestions.}
\label{tab:timequ_stats}
\centering
\small
\begin{tabular}{lccc}
\toprule
\textbf{Category} & \textbf{Train} & \textbf{Dev} & \textbf{Test} \\
\midrule
Explicit          & 2,724          & 1,302        & 1,311         \\
Implicit          & 651            & 291          & 292           \\
Temporal          & 2,657          & 1,073        & 1,067         \\
Ordinal           & 938            & 570          & 567           \\
\midrule
\textbf{Total}    & \textbf{6,970} & \textbf{3,236} & \textbf{3,237} \\
\bottomrule
\end{tabular}
\end{table}

\section{Experiments}
\subsection{Experimental Setup}

\stitle{Datasets.} We evaluate the effectiveness of TCC-RAG on two widely used TKGQA benchmarks: MULTITQ \cite{chen2023multi} and {TimeQuestions} \cite{sharma2023twirgcn}, as summarized in Table~\ref{tab:multiTQ_stats} and Table~\ref{tab:timequ_stats}.
MULTITQ is a large-scale TKGQA dataset containing over 500K questions with diverse temporal granularities. The questions span multiple temporal reasoning patterns, requiring varying numbers of reasoning steps over temporal knowledge graphs. Its scale and diversity make it particularly suitable for evaluating both retrieval accuracy and reasoning robustness under complex temporal settings. {TimeQuestions} is a complementary benchmark consisting of approximately 16K questions, categorized into four distinct temporal reasoning types. Across both datasets, different question types inherently require different levels of temporal reasoning, ranging from single-step temporal lookup to multi-step causal chain reasoning, making them well aligned with our evaluation objectives.

\begin{table*}[t]
\centering
\caption{Results of TCC-RAG and baselines on Hits@1 and Hits@10 on MULTITQ. The best and second best results are marked in {bold} and {underlined}, respectively. The bottom row is the relative improvement of TCC-RAG over the best baseline.}
\label{tab:mainresult_multitq}
\begin{tabular}{lcccccccccc} %
\toprule
\multirow{3}{*}{\textbf{Model}}
& \multicolumn{5}{c}{\textbf{Hits@1}}
& \multicolumn{5}{c}{\textbf{Hits@10}} \\
\cmidrule(lr){2-6} \cmidrule(lr){7-11}
& \textbf{Overall}
& \multicolumn{2}{c}{\textbf{Question Type}}
& \multicolumn{2}{c}{\textbf{Answer Type}}
& \textbf{Overall}
& \multicolumn{2}{c}{\textbf{Question Type}}
& \multicolumn{2}{c}{\textbf{Answer Type}} \\
\cmidrule(lr){3-4} \cmidrule(lr){5-6}
\cmidrule(lr){8-9} \cmidrule(lr){10-11}
& & \textbf{Multiple} & \textbf{Single} & \textbf{Entity} & \textbf{Time}
& & \textbf{Multiple} & \textbf{Single} & \textbf{Entity} & \textbf{Time} \\
\midrule
BERT          & 0.083 & 0.061 & 0.092 & 0.101 & 0.040 & 0.441 & 0.392 & 0.461 & 0.531 & 0.222 \\
DistilBERT    & 0.083 & 0.074 & 0.087 & 0.102 & 0.037 & 0.482 & 0.426 & 0.505 & 0.591 & 0.216 \\
ALBERT        & 0.108 & 0.086 & 0.116 & 0.139 & 0.032 & 0.484 & 0.415 & 0.512 & 0.589 & 0.228 \\
\midrule
EmbedKGQA     & 0.206 & 0.134 & 0.235 & 0.290 & 0.001 & 0.459 & 0.439 & 0.467 & 0.648 & 0.001 \\
CronKGQA      & 0.279 & 0.134 & 0.337 & 0.328 & 0.156 & 0.608 & {0.453} & 0.671 & 0.696 & 0.392 \\
MultiQA       & 0.293 & 0.159 & 0.347 & 0.349 & 0.157 & {0.635} & \underline{0.519} & {0.682} & \underline{0.733} & {0.396} \\
\midrule
LLaMA~3.1~8B        & 0.032 & 0.008 & 0.041 & 0.037 & 0.019 & 0.056 & 0.015 & 0.071 & 0.064 & 0.039 \\
GPT-4o mini        & 0.064 & 0.019 & 0.081 & 0.071 & 0.048 & 0.094 & 0.030 & 0.118 & 0.115 & 0.048 \\
DeepSeek-Chat      & 0.104 & 0.034 & 0.129 & 0.115 & 0.081 & 0.153 & 0.049 & 0.191 & 0.181 & 0.090 \\
ARI           & 0.380 & 0.210 & 0.680 & 0.394 & 0.344 & - & - & - & - & - \\
$\text{TimeR}^4$        & {0.728} & {0.335} & {0.887} & {0.639} & \underline{0.945} & - & - & - & - & - \\
RTQA          & {0.765} & {0.424} & \underline{0.902} & {0.692} & {0.942}
                 & {0.768}& 0.427 & {0.907} & {0.697} & \underline{0.942} \\
\midrule
{TCC-RAG}  \\ %
\enspace LLaMA~3.1~8B  & {0.762} & {0.580} & {0.836} & {0.715} &
                  {0.876}& {0.823} & {0.648} & {0.894} & {0.789} & {0.901} \\
\enspace GPT-4o mini & \underline{0.817} & \underline{0.646} & {0.879} & \underline{0.767} & {0.923}
                 & \underline{0.857}& \underline{0.709} & \underline{0.911} & \underline{0.821} &{0.936} \\
\enspace DeepSeek-Chat        & \textbf{0.847} & \textbf{0.691} & \textbf{0.911} & \textbf{0.801} & \textbf{0.959}
                & \textbf{0.876}& \textbf{0.746} & \textbf{0.928} & \textbf{0.839} & \textbf{0.961} \\ \midrule
\enspace Imp (\%)        & {10.72} & {62.97} & {1.00} & {15.75} & {1.48}
                & {14.06}& {43.74} & {2.32} & {14.46} & {2.02} \\

\bottomrule
\end{tabular}
\end{table*}

\stitle{Baselines.}
We compare TCC-RAG with a diverse set of baselines covering pretrained language models, embedding-based QA methods, and large language model (LLM)-based approaches.
On the {MULTITQ} dataset, we evaluate against the following methods:
\begin{itemize}[leftmargin=*]
    \item \textbf{Pretrained Language Models:} BERT \cite{devlin2019bert}, DistilBERT \cite{sanh2019distilbert}, and ALBERT \cite{Lan2020ALBERT}, which fine-tune pretrained transformers for temporal QA.
    \item \textbf{Embedding-based Methods:} EmbedKGQA \cite{saxena2020iEmbQA}, CronKGQA \cite{saxena2021cronQA}, and MultiQA \cite{chen2023multi}, which rely on learned knowledge graph embeddings for temporal reasoning.
    \item \textbf{LLM-based Methods:} LLaMA~3.1~8B \cite{grattafiori2024llama3}, GPT-4o-mini \cite{achiam2023gpt}, and DeepSeek-Chat \cite{liu2024deepseek} answer questions through prompt-based inference. ARI \cite{chen-2024-ari}, TimeR$^4$ \cite{qian2024timer4}, and RTQA \cite{gong2025rtqa} augment LLMs with semantic RAG pipelines.
\end{itemize}
On the {TimeQuestions} dataset, we compare against the following categories of baselines:
\begin{itemize}[leftmargin=*]
    \item \textbf{KGQA Methods:} PullNet~\cite{sun2019pullnet}, GRAFTNet~\cite{sun2018open}, and Uniqorn~\cite{pramanik2024uniqorn}, which perform reasoning over static or hybrid knowledge graphs. TKGQA-specific methods include CronKGQA \cite{saxena2021cronQA}, TempoQR~\cite{mavromatis2022tempoqr}, EXAQT~\cite{jia2021complex}, LGQA~\cite{liu2023local}, and TwiRGCN \cite{sharma2023twirgcn}.
    \item \textbf{LLM-based Methods:} LLaMA2~7B \cite{touvron2023llama}, ChatGPT \cite{achiam2023gpt}, GenTKGQA~\cite{gao2024two}, TimeR$^4$~\cite{qian2024timer4}, and RTQA~\cite{gong2025rtqa}, which leverage large language models with varying retrieval strategies.
\end{itemize}

These baselines collectively cover a wide spectrum of paradigms, enabling a comprehensive evaluation of TCC-RAG across different temporal reasoning and retrieval settings. Hits@$k$ measures the fraction of queries with a gold answer among the top-$k$ predictions in the ordered answer list ($k=1,10$).

\stitle{Implementation Details.}\label{app_imple}
\begin{itemize}[leftmargin=*]
    \item \textbf{LLM \& Model Configuration.} For the progressive query parsing module (Section~\ref{sec:retrieval}), we use a locally hosted LLaMA~3.1~8B \cite{grattafiori2024llama3} for both entity extraction and relation extraction. The NER component adopts {roberta-large-ner-english}\cite{liu2019roberta} as the surface-form recognition model, and {multi-qa-MiniLM-L6-cos-v1}\cite{wang2020minilm,reimers2019sentence} for cosine similarity computation between entity mentions and KG entities. For final answer generation, we evaluate TCC-RAG with three LLM backbones: DeepSeek-Chat~v3.1-20250922 \cite{liu2024deepseek}, GPT-4o-mini \cite{achiam2023gpt}, and LLaMA~3.1~8B \cite{grattafiori2024llama3}. On TimeQuestions, TCC-RAG is evaluated using DeepSeek-Chat~v3.1-20250922 as the generator. The entity-grounding similarity threshold is set to $\theta_{1}=0.85$. Candidates below this threshold fall back to fuzzy string matching. The temporal-bucket width $\delta$ in the causal signature $\kappa(\tau, \hat{c}, \tau_{b})$ is set to $1$~day for range cues (\texttt{before}, \texttt{after}). For \texttt{same}, \texttt{first}, and \texttt{last}, $\kappa$ uses exact-timestamp or ordinal-rank encoding without bucketing. CEF filters evidence only when its size-and-redundancy score exceeds a preset threshold, retaining it unchanged otherwise.

    \item \textbf{Baseline Setup.} On \textbf{MULTITQ}, results for non-LLM baselines (BERT/DistilBERT/ALBERT, EmbedKGQA, CronKGQA, MultiQA) and ARI/RTQA are taken directly from RTQA~\cite{gong2025rtqa}. TimeR$^{4}$ is evaluated using its publicly released checkpoint with its original pretrained LLaMA backbone. TCC-RAG accesses DeepSeek-Chat and GPT-4o-mini through public APIs, while LLaMA~3.1~8B is deployed locally. On \textbf{TimeQuestions}, results for non-LLM baselines (PullNet, Uniqorn, GRAFT-Net, CronKGQA, TempoQR, EXAQT, TwiRGCN, LGQA) are taken directly from TimeR$^{4}$~\cite{qian2024timer4}. For LLM-based RAG baselines (TimeR$^{4}$ and RTQA), we re-run them with the same frozen DeepSeek-Chat~v3.1-20250922 generator as TCC-RAG, rather than using their originally reported numbers. This is because the TimeQuestions training and test splits have partial overlap, which can cause data leakage when a fine-tuned backbone is used, inflating reported scores. By replacing all LLM generators with the same frozen DeepSeek-Chat~v3.1-20250922, we eliminate backbone confounds and ensure a fair comparison that isolates each method's retrieval and reasoning strategy.

    \item \textbf{Hardware \& Decoding.} All experiments are conducted on a server equipped with an NVIDIA A6000 GPU and an Intel Xeon Gold 6330 CPU @ 2.00GHz. All LLM calls use deterministic decoding with temperature $=0$ and top-$p=0.9$ to ensure reproducibility.
\end{itemize}

\subsection{Overall Performance}
\stitle{Results on MULTITQ.}
Table~\ref{tab:mainresult_multitq} reports the experimental results on the MULTITQ dataset. Overall, TCC-RAG consistently outperforms all baseline methods across evaluation metrics. In particular, compared with the state-of-the-art RTQA, TCC-RAG achieves relative improvements of 10.72\% and 14.06\% on Hits@1 and Hits@10, respectively. A closer inspection reveals that the performance gains are primarily driven by the \emph{Multiple} question category, where TCC-RAG achieves improvements of 62.97\% on Hits@1 and 74.71\% on Hits@10. This category exclusively consists of complex temporal questions requiring multi-step reasoning, such as \emph{Before\_Last}, \emph{After\_First}, and \emph{Equal Multi}. Existing semantic-based RAG methods struggle in this setting, as they are unable to retrieve coherent multi-hop causal evidence and often introduce redundant or irrelevant facts. In contrast, TCC-RAG explicitly leverages causal cue-guided traversal to retrieve multi-hop causal knowledge. By following temporal causal chains while suppressing redundant information, the proposed method is able to recover causally consistent evidence paths, leading to substantial improvements under complex multi-step temporal reasoning scenarios.

In addition, we further evaluate TCC-RAG with different LLM backbones. Notably, TCC-RAG equipped with LLaMA~3.1~8B achieves comparable Hits@1 performance to RTQA using DeepSeek-Chat, while surpassing it by 7.16\% on Hits@10. This result demonstrates that even with a lightweight backbone, TCC-RAG is able to compensate for limited reasoning capacity through effective causal retrieval.
When equipped with GPT-4o-mini, TCC-RAG further establishes clear advantages over RTQA, achieving improvements of 6.80\% and 11.58\% on Hits@1 and Hits@10, respectively. These consistent gains across different backbone capacities indicate that the performance improvements of TCC-RAG are not solely attributed to stronger language models, but rather stem from its superior ability to retrieve and organize temporally coherent causal knowledge.
By explicitly capturing intrinsic temporal causal structures through causal-guided retrieval, TCC-RAG effectively mitigates the reasoning limitations of smaller-scale LLMs and amplifies the strengths of more capable backbones, enabling it to outperform LLM-based baselines across model sizes consistently.

\begin{table}[t]
\centering
\caption{Performance comparison (Hits@1) on different question and answer types on TimeQuestions. The best is bolded and the second best is underlined. The bottom row is the relative improvement of TCC-RAG over the best baseline.}
\label{tab:timequestion}
\begin{adjustbox}{max width=\linewidth}
\begin{tabular}{lccccc}
\toprule
\multirow{2}{*}{\textbf{Model}} & \multirow{2}{*}{\textbf{Overall}} & \multicolumn{2}{c}{\textbf{Question Type}} & \multicolumn{2}{c}{\textbf{Answer Type}} \\
\cmidrule(lr){3-4} \cmidrule(lr){5-6}
& & Explicit & Implicit & Temporal & Ordinal \\
\midrule

PullNet & 0.105 & 0.022 & 0.081 & 0.234 & 0.029 \\
Uniqorn & 0.331 & 0.318 & 0.316 & 0.392 & 0.202 \\
GRAFT-Net & 0.452 & 0.445 & 0.428 & 0.515 & 0.322 \\
\midrule
CronKGQA & 0.462 & 0.466 & 0.445 & 0.511 & 0.369 \\
TempoQR & 0.416 & 0.465 & 0.036 & 0.400 & 0.349 \\
EXAQT & 0.572 & 0.568 & 0.512 & 0.642 & 0.420 \\
TwiRGCN & {0.605} & {0.602} & {0.586} & {0.641} & {0.518} \\
LGQA & 0.529 & 0.532 & 0.506 & 0.605 & 0.402 \\
\midrule
LLaMA2~7B & 0.271 & 0.268 & 0.325 & 0.279 & 0.234 \\
ChatGPT & 0.459 & 0.433 & 0.511 & 0.465 & 0.481 \\
GenTKGQA & 0.584 & 0.596 & \underline{0.611} & 0.563 & \underline{0.578} \\
$\text{TimeR}^4$ & {0.617} & {0.609} & {0.512} & {0.754} & {0.426} \\
RTQA & \underline{0.655} & \underline{0.648} & {0.601} & \underline{0.762} & {0.503} \\
\midrule
TCC-RAG & \textbf{0.710} & \textbf{0.711} & \textbf{0.694} & \textbf{0.781} & \textbf{0.597} \\
\midrule
Imp (\%) & {8.40} & {9.72} & {13.58} & {2.49} & {3.29} \\
\bottomrule
\end{tabular}
\end{adjustbox}
\end{table}

\stitle{Results on TimeQuestions.}
The experimental results on the {TimeQuestions} dataset are reported in Table~\ref{tab:timequestion}. As shown, {TCC-RAG} consistently achieves the best performance across all question types. Overall, {TCC-RAG} outperforms the best-performing baseline RTQA by 8.40\%, demonstrating its strong generalization ability on diverse temporal reasoning tasks.
Similar to the observations on MULTITQ, the performance gains on {TimeQuestions} are primarily attributed to question types that require multi-step temporal reasoning. In particular, on the \emph{Implicit} question category, {TCC-RAG} achieves a notable improvement of 13.58\% over the best baseline and 35.54\% over semantic-based RAG methods such as $\text{TimeR}^4$. Such questions require inferring multi-hop temporal and causal dependencies that are not explicitly stated in the query or directly retrievable via semantic similarity.
These results further indicate that {TCC-RAG} is able to enhance the reasoning capability of LLMs by explicitly retrieving multi-step temporal causal chains. By capturing structured causal knowledge rather than isolated facts, TCC-RAG effectively supports complex temporal reasoning, consistently validating its causal knowledge mining ability across different datasets and reasoning settings.

\subsection{Ablation Study}

We conduct ablation studies on MULTITQ to evaluate the effectiveness of each component in TCC-RAG. The results are reported in Table~\ref{tab:ablation result}. Specifically, \textbf{w/o CCG} removes the Causal Cue Graph and relies solely on semantic retrieval, \textbf{w/o PQP} disables the progressive query parsing pipeline and uses a single-shot LLM call to extract all signals (entities, relation, cue, role) at once, and \textbf{w/o CEF} removes the causal evidence filter.

\begin{table}[t]

\caption{Ablation studies (Hits@1) of TCC-RAG on MULTITQ.}
\label{tab:ablation result}
\centering
\small
\setlength{\tabcolsep}{4pt}
\begin{tabular}{lccccc}
\toprule
\textbf{Model} & \textbf{Overall} & \textbf{Multiple} & \textbf{Single} & \textbf{Entity} & \textbf{Time} \\
\midrule
\textbf{w/o CCG} & 0.647 & 0.127 & 0.837 & 0.517 & 0.933 \\
\textbf{w/o PQP} & 0.778 & 0.653 & 0.824 & 0.759 & 0.821 \\
\textbf{w/o CEF} & 0.841 & 0.664 & 0.910 & 0.791 & 0.958 \\
\textbf{TCC-RAG} & \textbf{0.847} & \textbf{0.691} & \textbf{0.911} & \textbf{0.801} & \textbf{0.959} \\
\bottomrule
\end{tabular}
\end{table}

\stitle{Effect of Causal Cue Graph.}
Removing the Causal Cue Graph results in a substantial performance degradation across all metrics. In particular, the overall performance drops by 23.61\%, with the most significant decline observed in the \emph{Multiple} question category, where the score decreases by 81.62\%. The \emph{Entity}-type answers also suffer a notable drop of 35.45\%, while the performance degradation on \emph{Single} and \emph{Time}-type questions is relatively minor (8.12\% and 2.71\%, respectively).
This phenomenon is expected, as \emph{Multiple} questions exclusively consist of complex multi-step temporal reasoning tasks. Similarly, \emph{Entity}-type answers tend to involve more multi-hop temporal dependencies than \emph{Time}-type answers. These results demonstrate that the causal cue graph is critical for modeling complex temporal causal structures, and its removal severely impairs the model’s ability to perform multi-step temporal reasoning.

\stitle{Effect of Progressive Query Parsing.}
Disabling progressive query parsing and relying on a single-shot LLM call leads to consistent performance degradation across all categories, with an overall drop of 8.14\%. This degradation is particularly evident in \emph{Multiple} (5.49\%) and \emph{Time}-type (14.3\%) questions.
The results indicate that time-sensitive queries exhibit high linguistic and structural diversity, making accurate entity and relation extraction challenging when all signals are produced in a single LLM pass. Without the staged decomposition into entity grounding, context-aware relation parsing, and rule-based cue extraction, errors accumulate and prevent effective grounding of queries onto the causal cue graph. Consequently, the model fails to fully exploit the structured causal retrieval mechanism, resulting in reduced performance.

\begin{figure}[t]
\centering
\includegraphics[width=1.0\linewidth]{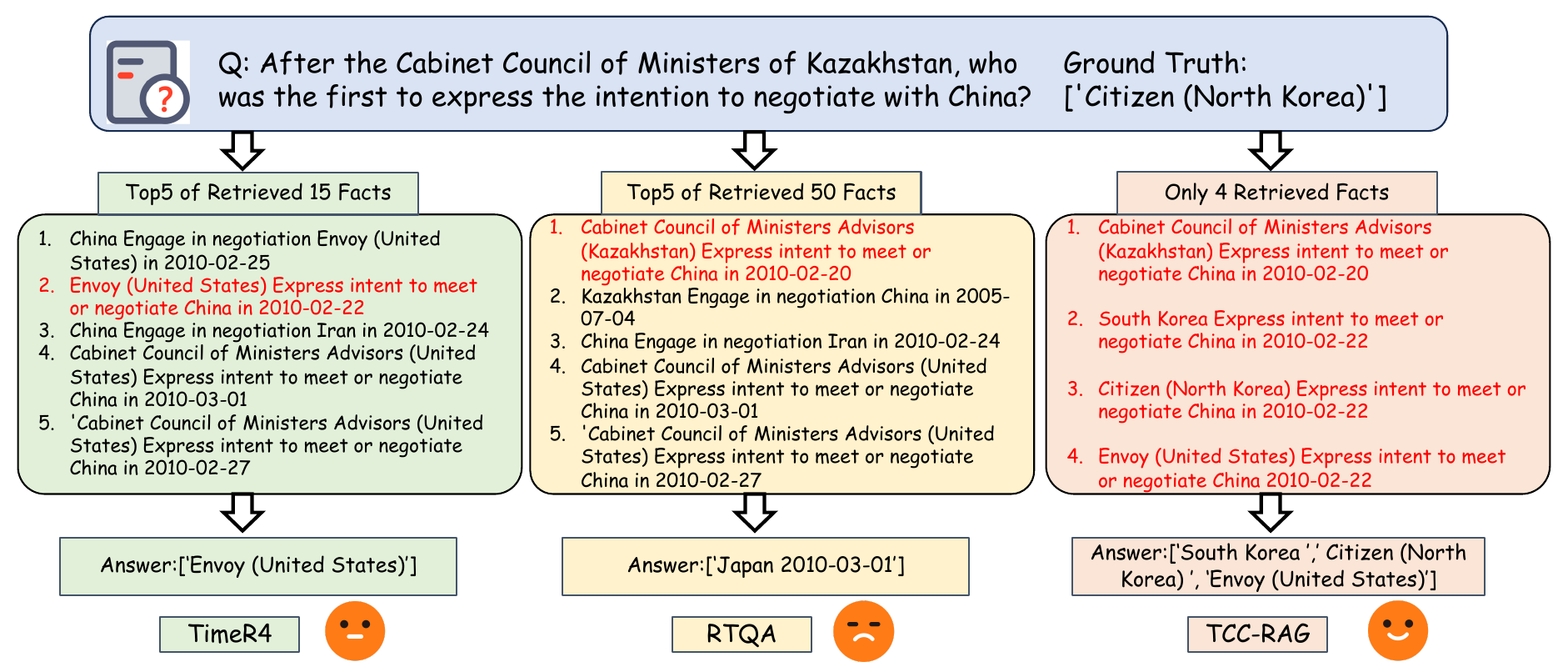}
\caption{A comparison of the retrieved facts between semantic-based RAG methods and TCC-RAG on MULTITQ. Facts highlighted in red are causally essential evidence.}

\label{fig_case}
\end{figure}

\stitle{Effect of Causal Evidence Filter.}
Removing the causal evidence filter also leads to performance degradation, particularly on \emph{Multiple} and \emph{Entity}-type questions, where the scores drop by 3.90\% and 1.24\%, respectively.
These observations suggest that the filter primarily contributes to refining causal evidence in complex temporal reasoning scenarios, where retrieval may still introduce redundant or noisy facts due to data noise. The relatively small overall performance drop (0.71\%) is expected, as the filter operates as a structural refinement layer on top of an already concise retrieval output. Since most queries can already retrieve minimal and precise causal knowledge through the causal cue graph, only a small portion of queries require further refinement by the filter.

\stitle{Semantic Fallback Analysis.}
Semantic retrieval is invoked when structured retrieval returns no facts. Table~\ref{tab:fallback_coverage} reports coverage on the complete MULTITQ test set and conditional Hits@1 within each retrieval path. Structured retrieval handles 46,102 of 54,584 queries, while fallback handles the remaining 8,482. With fallback disabled, uncovered queries receive no prediction, and overall Hits@1 is 0.770, still exceeding RTQA's 0.765. Thus, causal-cue retrieval is effective independently, while semantic fallback improves coverage.

\begin{table}[t]

  \caption{Retrieval-path coverage and conditional Hits@1 on MULTITQ.}
  \label{tab:fallback_coverage}
  \centering
  \small
  \begin{tabular}{llcc}
    \toprule
    \textbf{Dataset} & \textbf{Path} & \textbf{Coverage} & \textbf{Hits@1} \\
    \midrule
    \multirow{2}{*}{MULTITQ} & Structured & 84.46\% & 0.912 \\
                            & Fallback & 15.54\% & 0.492 \\
    \bottomrule
  \end{tabular}
\end{table}

\begin{figure*}[tbp]
\centering
\includegraphics[width=0.90\linewidth]{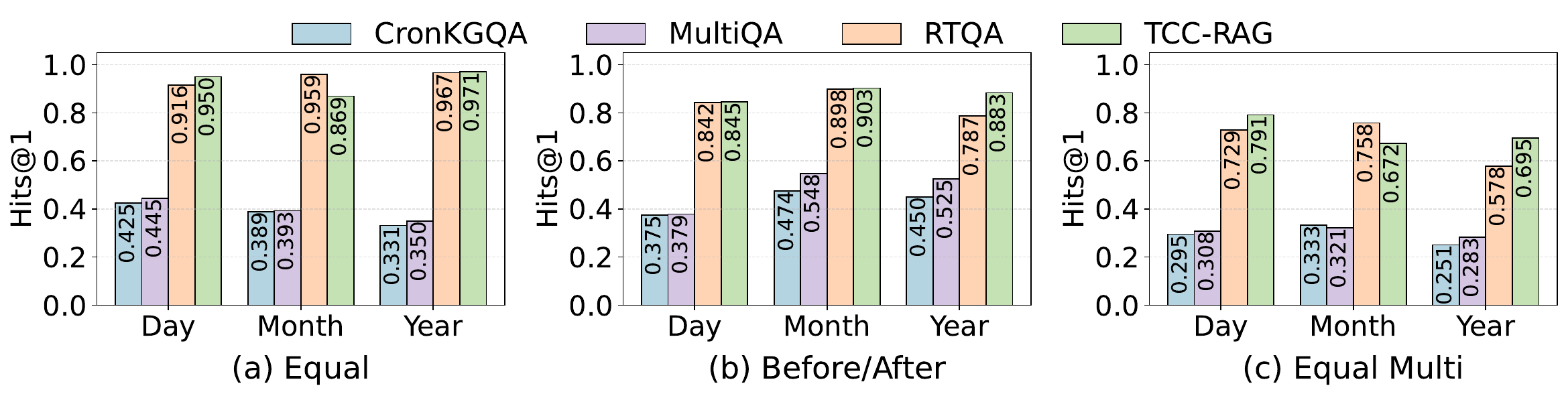}
\caption{Experiment results of Hits@1 across multiple temporal granularities on the MULTITQ dataset.}
\label{fig:time_gran}
\end{figure*}

\subsection{Insights and Case Study}

\stitle{Results for Multi-Granularity Time.} To evaluate the effectiveness of {TCC-RAG} under different temporal resolutions, we conduct experiments across multiple time granularities, including \emph{day}, \emph{month}, and \emph{year}. The results are shown in Figure~\ref{fig:time_gran}. Overall, TCC-RAG achieves the best performance across most temporal granularities and question types, demonstrating its robustness in handling temporal reasoning under different time resolutions.
In particular, consistent improvements are observed in the \emph{Before/After} and \emph{Equal Multi} categories, indicating that temporal causal chain-guided retrieval is effective for capturing causal temporal dependencies across events at different granularities.
We note that RTQA slightly outperforms TCC-RAG in the \emph{Equal-Month} and \emph{Equal-Multi-Month} settings. These cases mainly involve direct temporal alignment at a single reasoning step, where semantic retrieval is often sufficient. Nevertheless, TCC-RAG maintains stable performance across coarse to fine-grained temporal reasoning tasks, confirming the general effectiveness of the proposed causal retrieval framework.

\stitle{Case Study.}
To verify whether {TCC-RAG} can retrieve truly causal knowledge, we conduct a retrieval case study comparing {TCC-RAG} with semantic-based methods. The results are illustrated in Figure~\ref{fig_case}.
$\text{TimeR}^4$ employs rewrite-and-rerank strategies, while RTQA utilizes question decomposition. However, both methods fail to retrieve the key causal facts required for multi-step temporal reasoning and introduce a large number of redundant facts, which ultimately prevents the model from producing the completely correct answer.
{TCC-RAG} leverages the causal cue graph to retrieve only the most relevant facts. With merely four retrieved facts, the model can construct a complete temporal causal chain and successfully perform multi-step reasoning. This demonstrates that TCC-RAG is capable of mining {concise yet sufficient} causal knowledge.
Furthermore, this example reveals an implicit temporal ambiguity. Although the ground-truth answer is \emph{Citizen (North Korea)}, multiple events, including \emph{South Korea} and \emph{Envoy (United States)}, occur on the same day. TCC-RAG accurately captures this temporal coincidence through causal cue-guided retrieval and returns all temporally consistent answers. This observation further demonstrates that TCC-RAG can uncover latent causal structures within the temporal knowledge store rather than relying solely on surface-level semantic matching.

\begin{figure}[t]
\centering
\includegraphics[width=0.85\linewidth]{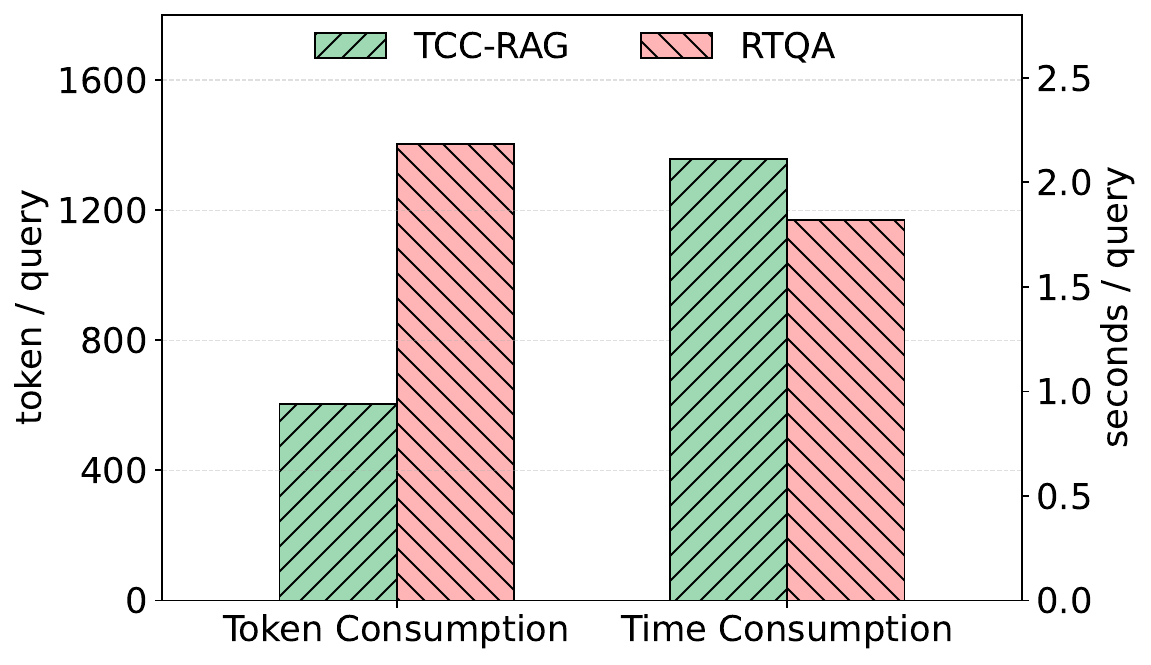}
\caption{Comparison of token cost and inference time between TCC-RAG and RTQA on MULTITQ.}
\label{fig:model_comparison}
\end{figure}

\subsection{Efficiency Analysis}\label{sec:efficiency}
Figure~\ref{fig:model_comparison} presents a detailed comparison of token cost and inference time between TCC-RAG and RTQA on MULTITQ, highlighting the efficiency advantages of TCC-RAG.

\stitle{Token Consumption.} We compare the average input token consumption of RTQA and TCC-RAG on the MULTITQ dataset, where both methods are deployed using DeepSeek-Chat. In RTQA, token usage mainly comes from the recursive question decomposition process and the subsequent answer aggregation for each sub-question. In contrast, TCC-RAG consumes tokens primarily during query parsing and final answer generation. Experimental results show that RTQA consumes an average of 1{,}403.8 tokens per question, whereas TCC-RAG only uses 604.8 tokens per question on average, reducing token consumption by over 57\%. This substantial reduction is mainly attributed to the concise yet sufficiently informative set of causal facts retrieved by TCC-RAG, which yields a much smaller evidence context for the LLM. These results demonstrate that causal-chain-guided retrieval not only improves reasoning accuracy but also significantly reduces token consumption compared with decomposition-based RAG pipelines.

\stitle{Inference Time.}
We further compare the average end-to-end processing time per query. TCC-RAG takes 2.11s per query on average, slightly higher than the 1.82s of RTQA. This minor gap mainly stems from the entity grounding stage, where TCC-RAG invokes both a NER model and an LLM for dual-path extraction and consistency verification, introducing a small additional cost. Nevertheless, this overhead is well justified by the substantial accuracy gains, and remains acceptable in practice. Since all LLM components are accessed through asynchronous API calls, multiple queries can be processed concurrently, allowing TCC-RAG to deliver answers efficiently at scale despite the slightly higher per-query latency.

\section{Conclusion}

In this paper, we propose {TCC-RAG}, a novel RAG framework for TKGQA. To overcome the limitations of existing semantic-based RAG methods, which suffer from missing causally relevant facts and introducing noisy evidence, we introduce a multi-view causal cue graph that explicitly encodes temporal causal relations. Built on this structure, we design a progressive causal chain-guided retrieval strategy that traces causally relevant evidence along the routed graph view, paired with a causal evidence filter that distills the retrieved set into a minimal causally sufficient subset. Extensive experiments on two widely used TKGQA benchmarks demonstrate that TCC-RAG consistently outperforms state-of-the-art methods, with especially significant improvements on complex multi-step temporal reasoning. Ablation studies verify the effectiveness of each component, while a case study further demonstrates the ability of TCC-RAG to retrieve concise yet sufficient causal evidence. Moreover, efficiency analysis confirms the practical efficiency and scalability of the proposed framework.

\section*{Acknowledgments}
This work was sponsored by the National Natural Science Foundation of China (No.62472327, 62572361), CCF-Tencent Rhino-Bird Open Research Fund (CCF-Tencent RAGR20260132), and the Sichuan Clinical Research Center for Imaging Medicine (YXYX2402).

\section*{AI-Generated Content Acknowledgments}
AI tools (ChatGPT and Claude) were used for language polishing and to suggest improvements to the visual presentation of selected figures, including colors and formatting. AI was not used for research conception, method design, or experiments.

\bibliographystyle{IEEEtran}
\bibliography{ref}

\end{document}